\documentclass[conference]{IEEEtran}  %

\usepackage[utf8]{inputenc}

\usepackage[american]{babel} %
\usepackage[usenames,dvipsnames]{xcolor}
\usepackage{cite}
\usepackage{float}
\usepackage{listings}
\usepackage{microtype}
\usepackage{booktabs}
\usepackage{url}
\usepackage{graphicx}
\usepackage{todonotes}
\usepackage[font=small,labelfont=bf]{caption}
\usepackage{subcaption}
\usepackage{gensymb}
\usepackage{array,multirow}
\usepackage{amssymb}
\usepackage{amsmath}
\usepackage{hyperref}
\usepackage{lipsum}
\usepackage{multicol}
\usepackage{subcaption}
\usepackage{siunitx}

\addto\extrasamerican{%

}
\usepackage{balance}

\DeclareSIUnit{\fps}{ \translate{fps} }

\makeatletter
\newcommand*\titleheader[1]{\gdef\@titleheader{#1}}
\AtBeginDocument{%
  \let\st@red@title\@title
  \def\@title{%
    \bgroup\color{gray}\raggedright\vspace{-1.7cm}\small\@titleheader\par\egroup
    \vskip1.5em\centering\normalfont\st@red@title}
}
\makeatother

\title{Mesh-based Object Tracking for Dynamic Semantic 3D Scene Graphs via Ray Tracing}
\titleheader{Accepted to RSS Workshop on Semantics for Robotics: From Environment Understanding and Reasoning to Safe Interaction 2024}

\author{
    \IEEEauthorblockN{Lennart Niecksch\IEEEauthorrefmark{1}, Alexander Mock\IEEEauthorrefmark{2}, Felix Igelbrink\IEEEauthorrefmark{1}, Thomas Wiemann\IEEEauthorrefmark{3}, Joachim Hertzberg\IEEEauthorrefmark{1}\IEEEauthorrefmark{2}}
    \IEEEauthorblockA{\IEEEauthorrefmark{1}German Research Centre for Artificial Intelligence \\ Plan-based Robot Control, Osnabrück, Germany\\
    Email: \{firstname.lastname\}@dfki.de}
    \IEEEauthorblockA{\IEEEauthorrefmark{2}Osnabrück University, Institute of Computer Science\\ Knowledge-based Systems, Osnabrück, Germany\\
    Email: \{amock, jhertzberg\}@uos.de} 
    \IEEEauthorblockA{\IEEEauthorrefmark{3}Fulda University of Applied Sciences, Department of Applied Computer Science \\
    Robotics in Computer Science, Fulda, Germany\\
    Email: thomas.wiemann@informatik.hs-fulda.de}
}

\begin{document}

\setlength{\abovedisplayskip}{3pt}
\setlength{\belowdisplayskip}{3pt}

\maketitle
\thispagestyle{empty}
\pagestyle{empty}

\begin{abstract}

In this paper, we present a novel method for 3D geometric scene graph generation using range sensors and RGB cameras.
We initially detect instance-wise keypoints with a YOLOv8s model to compute 6D pose estimates of known objects by solving PnP.
We use a ray tracing approach to track a geometric scene graph consisting of mesh models of object instances.
In contrast to classical point-to-point matching, this leads to more robust results, especially under occlusions between objects instances.
We show that using this hybrid strategy leads to robust self-localization, pre-segmentation of the range sensor data and accurate pose tracking of objects using the same environmental representation.
All detected objects are integrated into a semantic scene graph.
This scene graph then serves as a front end to a semantic mapping framework to allow spatial reasoning.

\end{abstract}

\section{Introduction}

In robotics, semantic 3D scene graphs can be used for navigation and interaction within an environment by helping robots understand and interact with their surroundings in a more human-like manner. 
Such scene graphs also allow for structured reasoning about the contained entities~\cite{armeni3DSceneGraph2019}.
Constructing scene graphs from images or videos involves significant challenges, including accurately detecting objects, determining their attributes, and identifying the complex relationships between them under various conditions. 
Advances in deep learning, especially in areas like object detection and relation prediction, have significantly contributed to progress in generating more accurate and detailed semantic 
scene graphs.
Tracking of the geometry and poses of known object instances over time is the most important step to create and update such scene graphs and connect the observations with spatial and background knowledge to allow reasoning.

In this paper, we present an approach to track the poses and spatial relations of object instances of known classes to create and update a geometric 3D scene graph.
For object detection, we use a YOLOv8~\cite{YoloV8} model that was trained to detect objects and 3D bounding box keypoints in camera data.
These rough pose estimates are refined by matching 3D reference mesh models to range sensor data.
For that, we use a ray-tracing approach to find correspondences between the meshes and the actual sensor data.
The detected objects are inserted into the scene graph and tracked over time.
In addition to building and updating the scene graph, we add the detected instances into our semantic mapping framework SEMAP~\cite{deeken2015semap} to automatically determine spatio-semantic relations such as \emph{on}, \emph{left-of} etc.
In preliminary experiments, we present first results to evaluate our approach in a real-world application example on a modified PAL Robotics Tiago robot.

\section{Related Work}

Research in 6D pose estimation of objects has seen significant advancements, leveraging various techniques from classical computer vision to deep learning \cite{tremblay2018corl:dope, kehl2016deep, wang2020self6d, wu2020eao, su2022zebrapose}, while other work focuses on instance segmentation in 3D scenes \cite{Dai_2018_ECCV, dai2018scancomplete, hou20193d}.
Generating scene graphs is typically accomplished using comprehensive scene information in the form of dense point cloud or mesh data, often obtained from simultaneous localization and mapping (SLAM) techniques \cite{rozenberszki2024unscene3d, wald2020learning, rosinol2020kimera, grinvaldVolumetricInstanceAwareSemantic2019}.

Kimera~\cite{rosinol2020kimera} is an approach to for real-time (SLAM) and 3D scene graph semantic reconstruction tailored for RGB-D data.
It focuses on creating a semantic scene graph of static objects of pre-known classes based on a TSDF representation of the environment. 
Building up on the Kimera scene graph, Hughes et al.~\cite{hughes2022hydra, hughes2023foundations} developed Hydra, a system that is able to segment, insert and track objects in the scene graph. 
This scene graph is a layered representation of spatial concepts at different levels of abstraction (i.e. places, rooms, buildings) and thus allows for further spatial reasoning, such as deriving the room types from the objects therein as presented in~\cite{strader2024indoor}.
However, this scene graph is still derived from a static map and while it is updated to changes occurring during the mapping process (i.e. loop closures), it is unable to adapt the changes once the mapping is completed and thus cannot track dynamic objects nor react to changes made during operation such as robotic manipulation.

Voxblox++~\cite{grinvaldVolumetricInstanceAwareSemantic2019} provides an instance-aware semantic mapping, which segments the scene into individual instances of semantic objects detected by a Mask R-CNN network. 
In contrast to Kimera, the objects are treated as individual geometries and thus can be tracked individually in the scene.
Because Voxblox++ does compute a semantic scene graph, it cannot be used for further spatial reasoning on the detected instances.

In~\cite{DEEKEN2018146}, Deeken et al. introduced SEMAP as a semantic mapping framework to model and infer spatial relations between objects and to reason about them.
This framework was successfully used in practical applications, e.g., to infer the spatial relations of moving machines in a maize harvesting campaign~\cite{deeken2019spatio}. 
However, inferring and updating the relations in this work is not real-time capable.

3D meshes are a standard data structure to represent real-world objects.
In contrast to Voxel/TSDF-based representations they are memory-efficient and CAD models of a broad range of common world objects are available or can be reconstructed using 3D scanners.
Dedicated hardware, such as NVIDIA's RTX make 3D meshes well-suited for real-time computation of spatial operations, such as ray intersection tests.
Consequently, they have recently been adopted as a map format for various robotics applications, including SLAM~\cite{vizzo2021puma, lin2023immesh, ruan2023slamesh}, map-based localization~\cite{mock2023micpl}, and path planning for navigation~\cite{puetz2021cvp}.
Integrating multiple meshes within a geometric scene graph enables an accurate representation of a real-world scene including multiple objects.

In this work, we demonstrate a system that shares the same geometric scene graph consisting of meshes for both map-based robot localization as well as dynamic object pose tracking.
Our main contribution is the scene graph creation with dynamic objects using RGB images as priors and Mesh-ICP~\cite{mock2023micpl} to refine their poses with depth sensor measurements.
The tracked 6D object are fed into SEMAP to connect the geometric information with semantic background knowledge.
Our proof-of-concept shows that this approach enables spatio-semantic reasoning in dynamic scenarios with moving objects, in our case a workbench scenario, where humans and robots interact.

\section{Scene Graph Generation}

\subsection{System Setup}
\begin{figure}[t]
    \centering
    \includegraphics[width=0.8\linewidth]{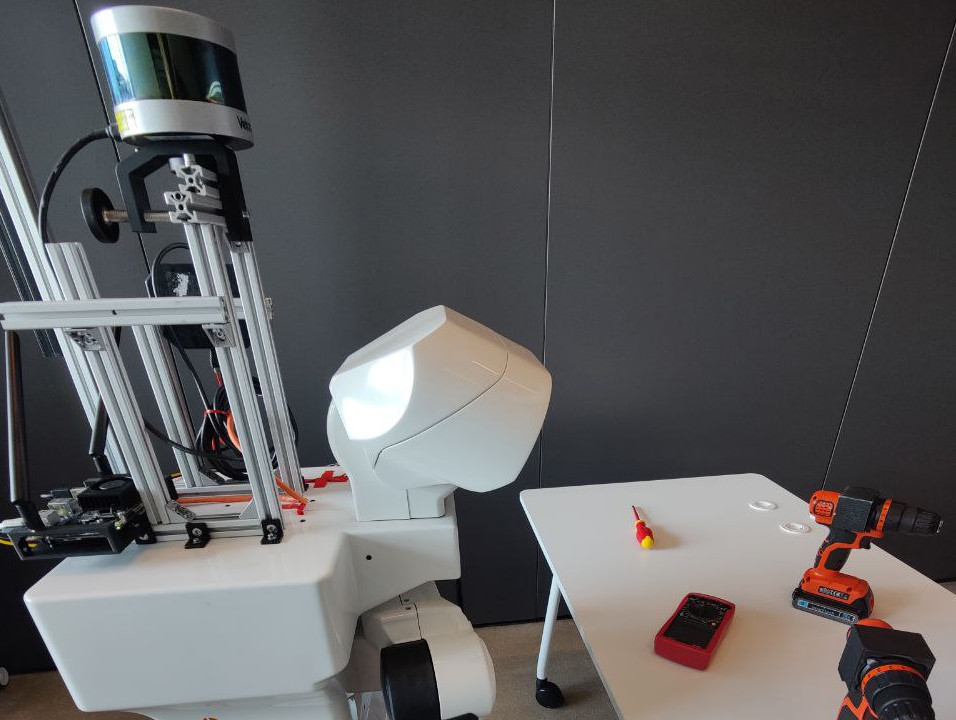}
    \caption{Our test platform Tiago inspecting a table top scene containing multiple objects.}
    \label{fig:tiago}
    \vspace{-0.5cm}
\end{figure}

Our test platform is a Tiago robot by PAL-robotics equipped with a ASUS Xtion camera in the head that produces RGB-D images at 30 fps and 640x480 pixel resolution (see \autoref{fig:tiago}).
We extended the stock setup with a NVIDIA Jetson Orin NX and a Velodyne VLP-16.
The Jetson operates with Ubuntu 20, Jetpack 5 and ROS1 Noetic to ensure message compatibility with Tiago's base.
Despite this old software stack, all developed software modules are compatible with ROS2.

To build a map of our lab environment we used a high-resolution laser scanner to capture the static parts of the environment.
We then reconstructed triangle meshes from the captured point cloud using LVR2~\cite{wiemann2018lvr}.

The static parts of the scene, i.e. the parts of the building, that are used for 3D self-localization using MICP-L~\cite{mock2023micpl} can be seen in the bottom picture in~\autoref{fig:presegment}.
MICP-L is a part of the RMCL library, which internally uses Rmagine~\cite{mock2023rmagine}, an optimized library for spatial operations in 3D geometric scene graphs, such as nearest neighbor searches or ray intersection tests.
Rmagine implements all functionality to create and update such 3D geometric scene graphs in real-time.
Rmagine has a strong focus on robotic applications and therefore offers interfaces to different CPU and GPU-based acceleration techniques.
It currently implements two backends: Embree~\cite{embree14} for CPUs and OptiX~\cite{optix10} for Nvidia GPUs - optionally accelerated by RTX units.

\begin{figure*}
 \begin{subfigure}{.323\textwidth}
    \centering
        \includegraphics[trim={0 50 0 50},clip,width=1.0\linewidth]{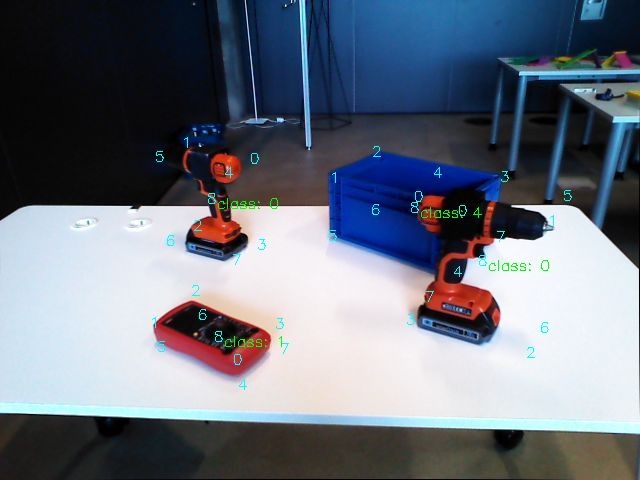}
        \caption{}
        \label{fig:keypoints}
    \end{subfigure}
    \begin{subfigure}{.323\textwidth}
        \centering
        \includegraphics[trim={0 50 0 50},clip,width=1.0\linewidth]{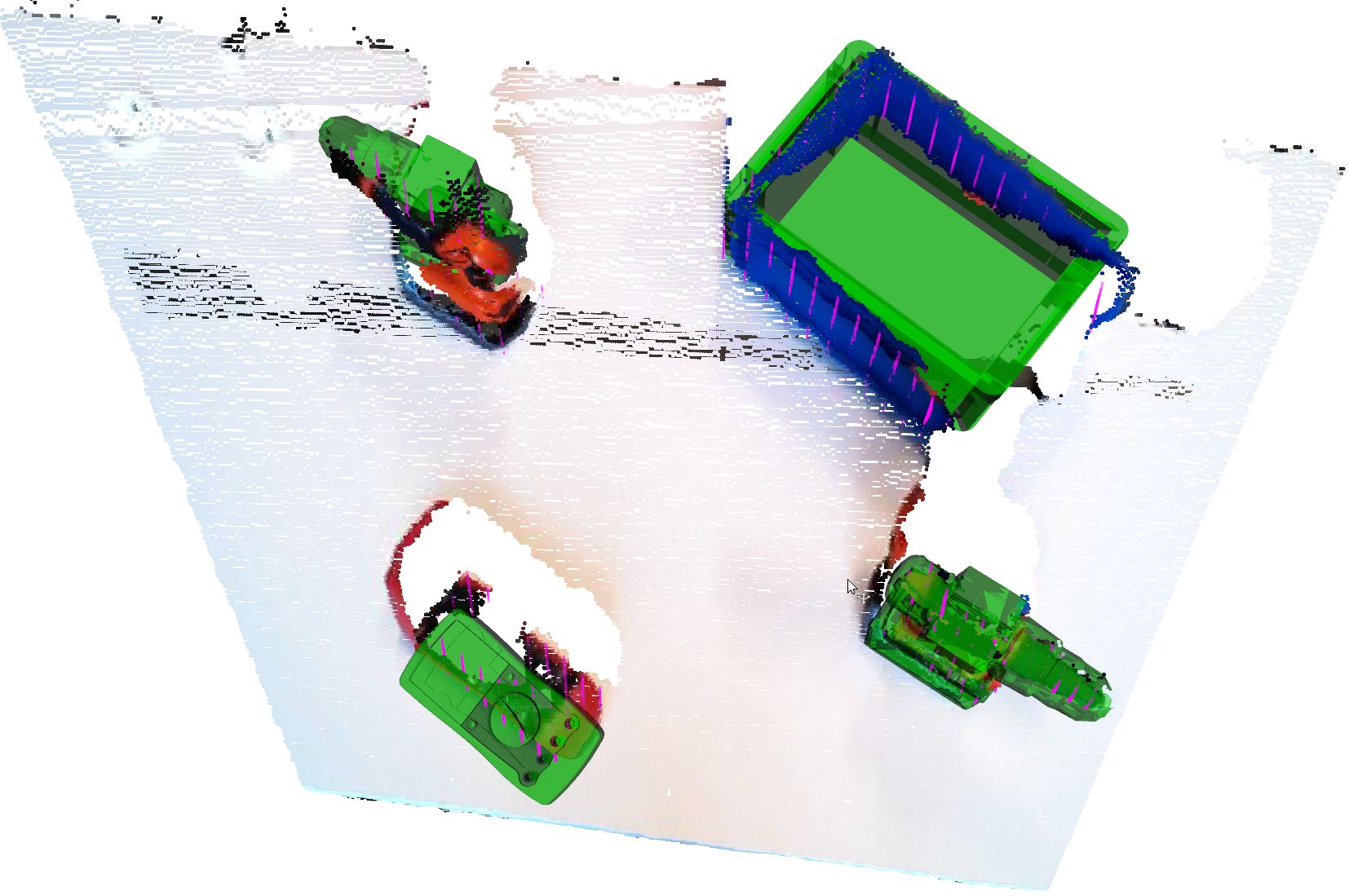}
        \caption{}
        \label{fig:correspondences}
    \end{subfigure}
     \begin{subfigure}{.323\textwidth}
        \centering
        \includegraphics[trim={0 50 0 50},clip,width=1.0\linewidth]{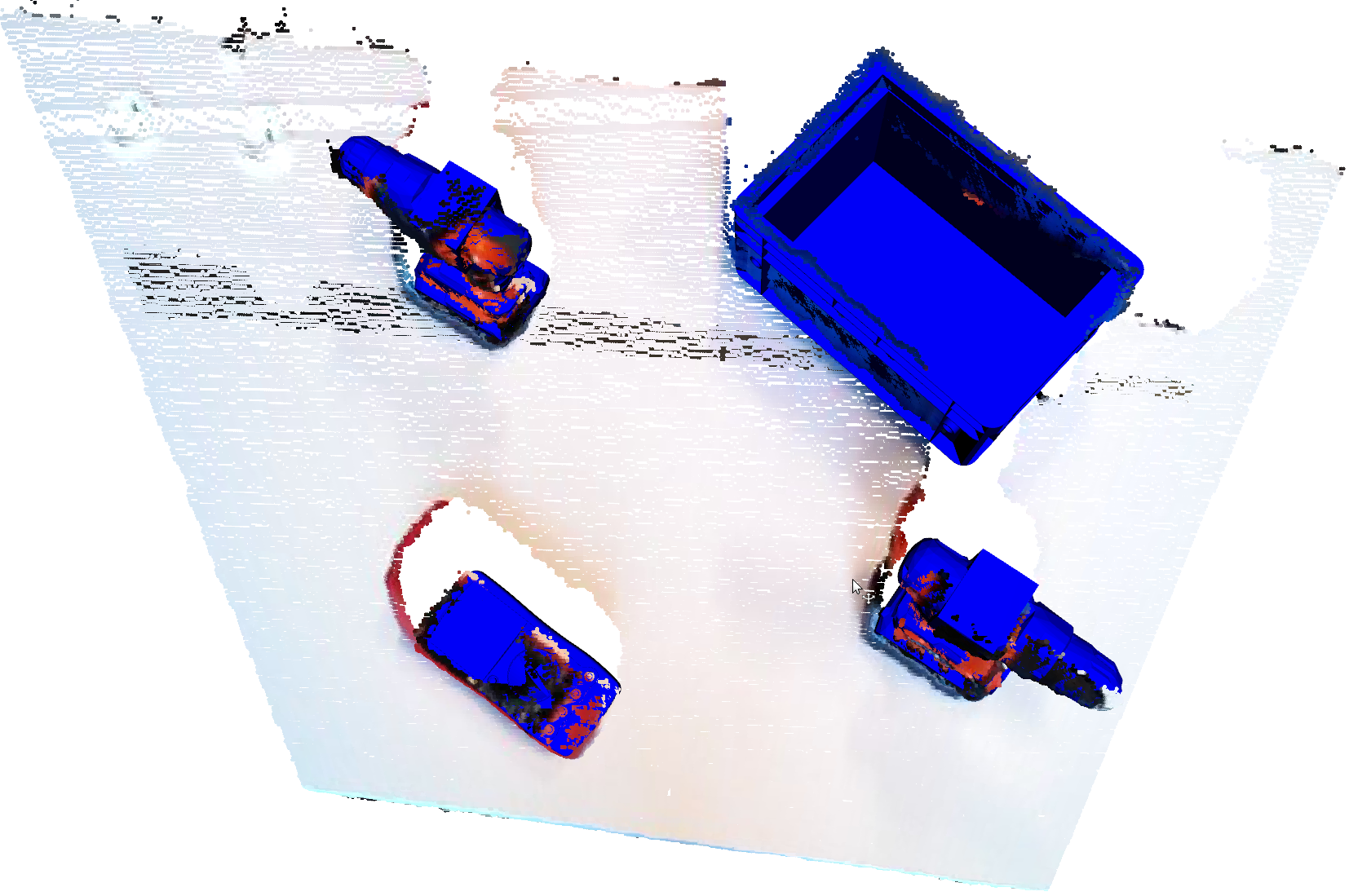}
        \caption{}
        \label{fig:corrected}
    \end{subfigure}
  \centering
  \caption{\subref{fig:keypoints} The scene and the detected keypoints and boxes. \subref{fig:correspondences} The initial pose estimates (green) and the ray tracing correspondences (purple). \subref{fig:corrected} The refined poses of the object instances (blue).}
  \label{fig:registration}
  \vspace{-0.5cm}
\end{figure*}

\subsection{Object Detection and Tracking}

Our approach for the initial object pose estimation is similar to \cite{tremblay2018corl:dope}.
First, the corners and centers of the detected bounding boxes are predicted in image space (see \autoref{fig:keypoints}).
The 6D pose is then estimated by minimizing the reprojection error of the known correspondences and dimension of the object bounding boxes using perspective-n-point (PnP). 
For fast prediction of the corners and centre points of object bounding boxes, we use a YOLOv8s~\cite{YoloV8} keypoint model trained on 54000 synthetic images generated with NVISII~\cite{morrical2021nvisii}.
The objects and their 3D meshes are described in detail in~\cite{lima2023physics}.

As the initial pose estimate solely relies on 2D image data, errors -- especially in depth and slight rotational ones -- are likely. Similar to other methods that use ICP on depth data to refine the initial guess, we determine correspondences for pose correction. Instead of determining them through point-to-point matching, e.g., by sampling the surface of the original models, we incorporate the sensor model and the known geometries directly.

The initial pose and the geometric priors are used to build an Embree scene graph~\cite{embree14} with Rmagine~\cite{mock2023rmagine}, containing individual geometric instances for each object, allowing to efficiently simulate the sensor using ray tracing. With the exact camera model of the depth camera, correspondences between the simulated and real sensor data are determined along the direction of the of a hitting ray as shown in~\autoref{fig:correspondences}.
Using the proper sensor model, the pose and the known geometries (meshes) of detected objects, naturally results in robust correspondences with respect to the previously described errors.
Additionally, the method accounts for (partial) occlusion between different instances by design.

As a result, each sensor data point $D_i(P)$ is pointing at a corresponding model point $M_i(P)$, referencing its surface normal $M_i(P)$, and an instance/object id $M_i(O)$ with $i \in [0, n]$, or in short $D_i(P) \rightarrow M_i(P, N, O)$.
We transform this into a set of correspondences per object instance: $O_i: D(P_i) \rightarrow M(P_i, N_i, C_i)$.
We discard correspondences based on a distance threshold which not only removes false assignments around the boundary, it additionally enables us to effectively prune false positives from the model.
The distance between each set of correspondences is individually minimized using~\cite{umeyama1991least}, resulting in a refinement for the initial guesses, which are used to update the instances transformation in the scene graph.
This intrinsically minimizes the distances between the 3D models and the sensor data and achieves high overall accuracy as shown in~\autoref{fig:registration}.
The procedure is then repeated.
To better account for uncertainty and occlusions, we plan to integrate a Kalman Filter per tracked object in future work.
This also helps when the object detector fails, because the information, that the simulation step still led to valid correspondences based on the filter's hypothesis, is integrated as well.
Inversely, tracks may be safely pruned with a high probability if the latter failed and no occlusion is given.

Besides being robust, our method is also computationally efficient.
With the initial pose guess of the robot, we start to find correspondences within a large scene.
As described in MICP-L~\cite{mock2023micpl}, this has a run time complexity of $O(n \cdot \log(m))$ with $n$ being the number of measurements considered and $m$ being a general parameter for the map's size, which can be either the number of primitives in a mesh or the number of meshes with the same number of faces.
Since the number of measurements per scan is constant, the overall run time only changes logarithmically with the map size.

Splitting the correspondences per object and computing the transformation parameters is linear w.r.t. the number of measurements, therefore it is not directly increasing with larger maps.
The actual update of the scene depends on how many objects are found with the correspondence search and thus depends on the situation.
Assuming we have a large scene with 100 rooms, each room consists of 10 objects to be tracked; 1000 objects in total.
The robot determines the correspondences with the scene, corrects the robot's pose, and starts to track the visible objects; all with a $O(\log(m))$ complexity.
Since the correspondences yield only the 10 objects placed in the same room as the robot, the problem is reduced to track 10 objects out of 1000.
If we escalate this example to even larger worlds, e.g., having 1 million rooms, the number of tracked objects remains 10.
Only the correspondence finding procedure increases, however only with a complexity of $O(\log(m))$.

\subsection{Sensor data pre-segmentation}

\begin{figure}[t]
    \centering
    \begin{subfigure}{.85\linewidth}
        \centering
        \includegraphics[trim={0 0 0 200},clip,width=1.0\linewidth]{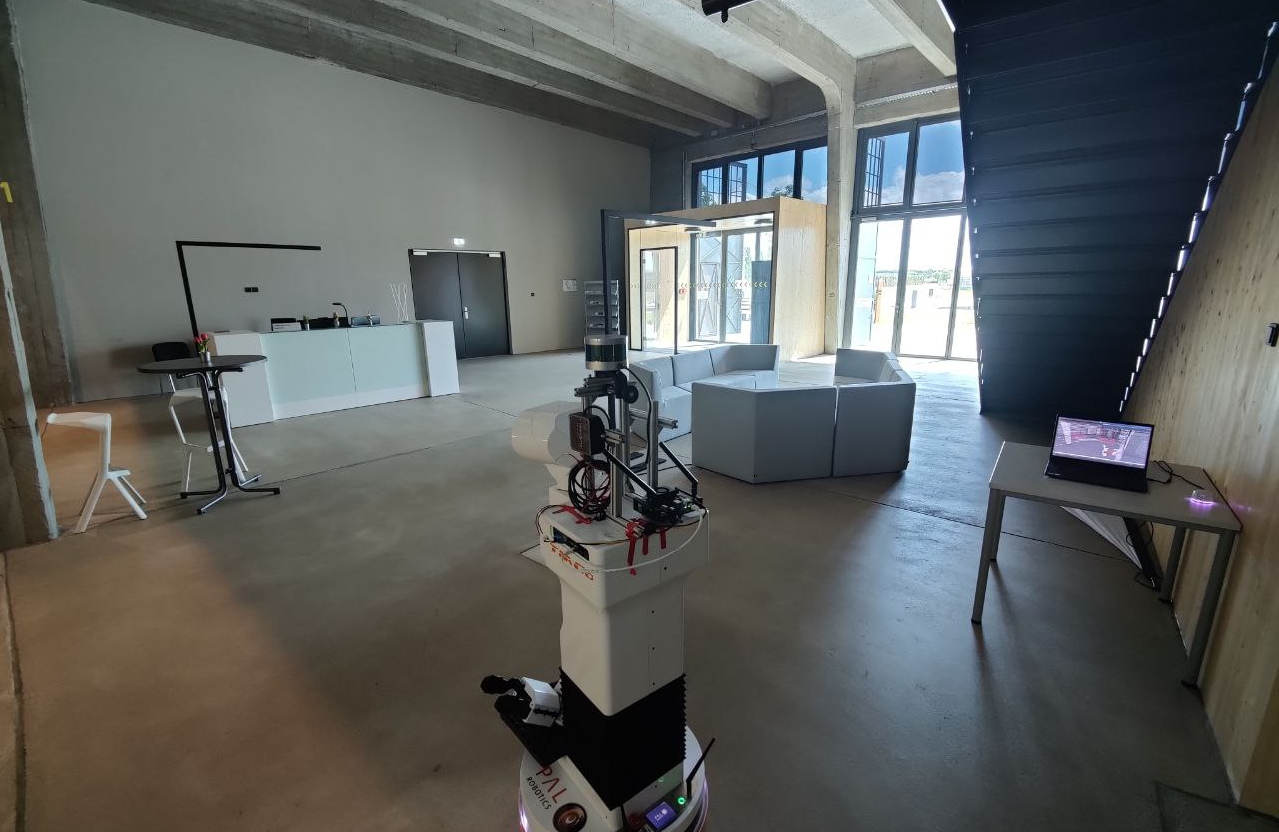}
    \end{subfigure}
    \begin{subfigure}{.85\linewidth}
        \centering
        \includegraphics[trim={0 100 0 60},clip,width=1.0\linewidth]{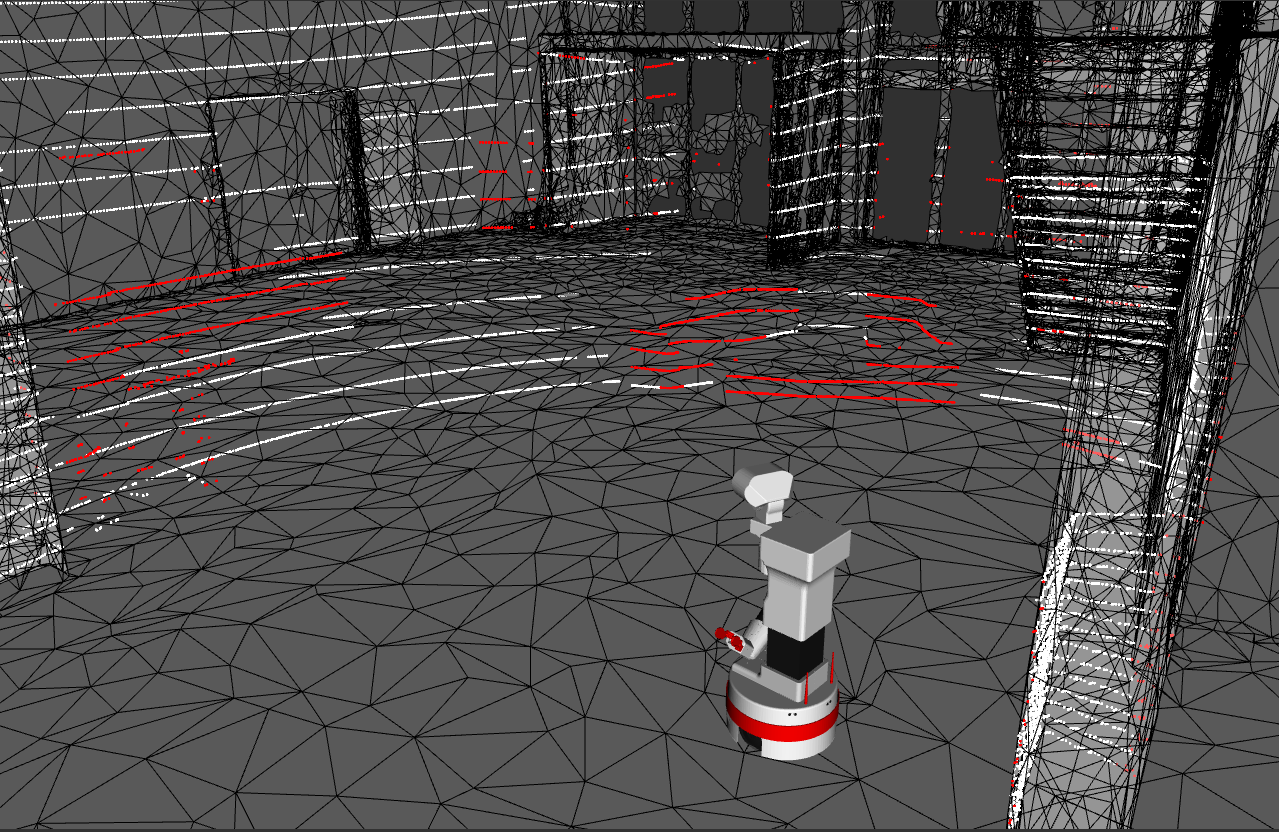}
    \end{subfigure}
    \caption{By tracking the geometric scene graph we can inversely determine all the points that are unknown and use it as pre-segmentation for other methods. The left image shows the Tiago robot operating inside a real environment with a matched scene graph in the right image, which is only composed of walls and doors.
    The points that are not considered for tracking (red) give a pre-segmentation of the sensor data that can help subsequent object detectors to produce more reliable results.
    }
    \label{fig:presegment}
    \vspace{-0.3cm}
\end{figure}

While we track parts of the scene graph using sets of correspondences, some measurements remain unassigned to any object, e.g., if a certain distance threshold is exceeded.
Those uncategorized measurements occur if an object is visible in the sensor data that is neither detected by YOLO nor is existing anywhere nearby in the current scene.
Such a situation is visualized in~\autoref{fig:presegment}.
This pre-segmentation of the unknown helps subsequent object detectors to recognize objects faster and more reliably, as the number of measurements is reduced to a minimum.
In the scene of \autoref{fig:presegment} we were able to reduce an Velodyne VLP-16 scan from 25287 valid measurements to 3515 points, i.e., a reduction of $\sim$86\% without losing relevant information at all.

\subsection{Integration with qualitative spatial reasoning}

\begin{figure}[t]
    \centering
    \begin{subfigure}{.49\linewidth}
        \centering
        \includegraphics[trim={4cm 0.5cm 5cm 1cm},clip,width=1.0\linewidth]{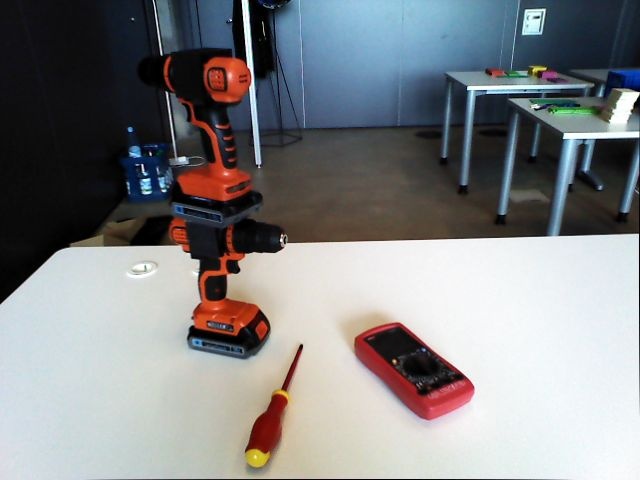}
    \end{subfigure}
    \begin{subfigure}{.49\linewidth}
        \centering
        \includegraphics[trim={32cm 6.5cm 22cm 3cm},clip,width=1.0\linewidth]{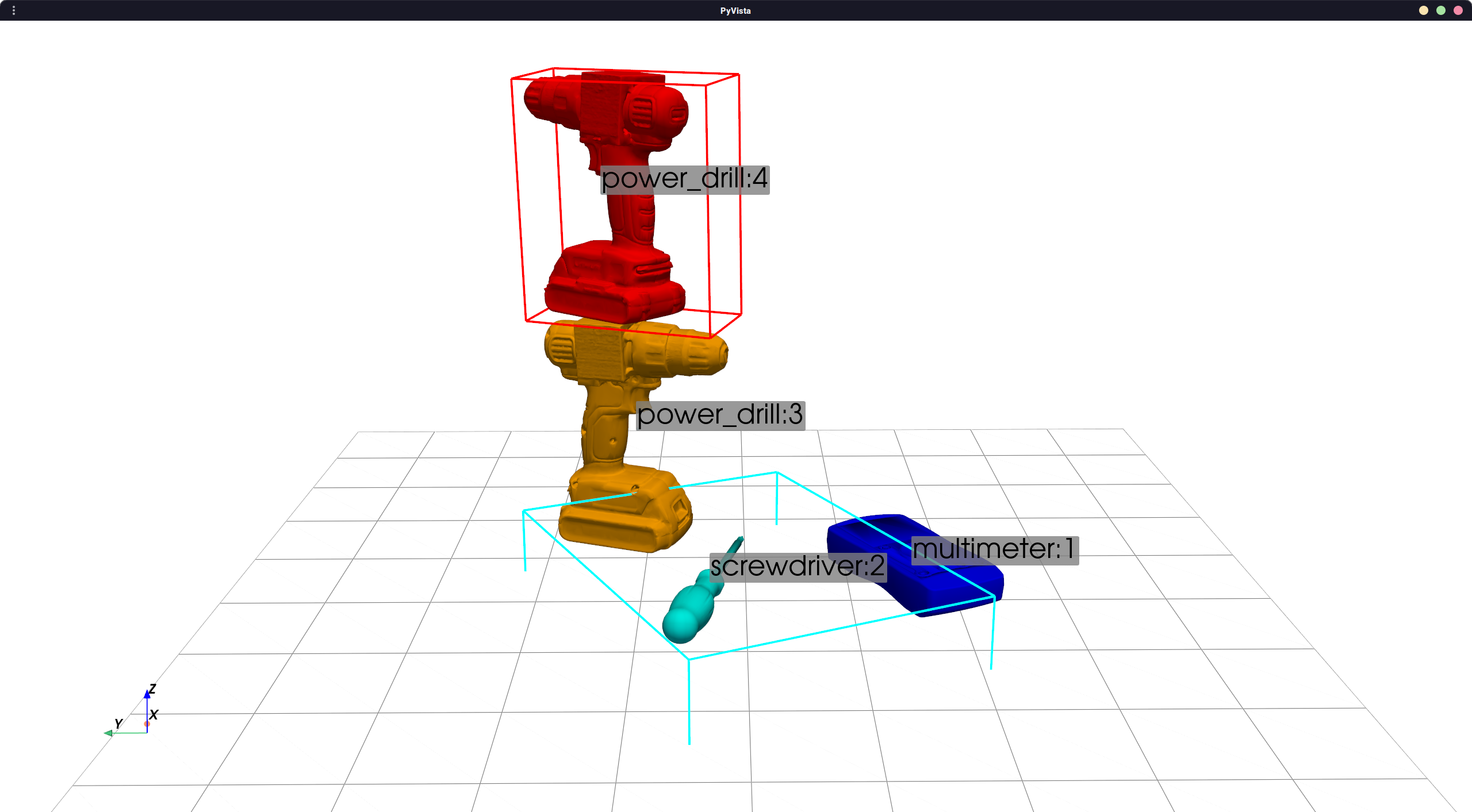}    
    \end{subfigure}
    \caption{Result of SEMAP queries \emph{on} (red) and \emph{left-of} (cyan) using the refined object poses. Best viewed in color. The red box shows the top projection abstraction of the bottom powerdrill and the cyan box the left projection abstraction of the multimeter.}
    \label{fig:semap}
    \vspace{-0.5cm}
\end{figure}

As a proof of concept, we input the recognized and tracked objects into SEMAP~\cite{deeken2015semap} for instance-based spatial reasoning.
\autoref{fig:semap} shows the result of a query for objects that are on top or left of other objects with respect to their local coordinate systems. Such information can be used to build hierarchical semantic scene graphs, or can directly be input into symbolic task planning algorithms.
In future work, we plan to use the same scene graph structure used for pose refinement for the derivation of topological relations in both the tracking front end and the semantic backend.
This would have several benefits.
First, the graph structure is already continuously updated in the refinement steps.
Second, the underlying acceleration structures allow for distance and collision queries and have already proven to be suitable for live updates and computations on full resolution mesh geometries.
Additionally, Rmagine allows for perspective adjustment, e.g., to clarify directional spatial queries.
For example, without semantics, a cup (c) that is standing in a cupboard (b) is just a cup and a cupboard, both with transforms relative to the scene root (m): $(T^{c}_{m}, T^{t}_{m})$. 
Now, if there was the actual semantic relation that the cup stands in the cupboard, the transformations/connections in the Rmagine scene can be updated to $(T^{c}_{t}, T^{t}_{m})$ by using all the already existing information $T^{c}_{t} = T^{t-1}_{m} \cdot T^{c}_{m}$. 
Mathematically, both sets of transformations represent the same.
These improvements are the foundation for a continuous monitoring of 3D spatial relations. 
This would enable the possibility of live 3D spatial and temporal semantic reasoning, e.g., via recurrent neural networks or pattern based complex event processing like~\cite{nieksch2023detecting}. 

\section{Preliminary Experiments}

Since we want to run most of the computations on edge and ensure real-time operation, it is important to keep the run times low.
Therefore, we measured the run times for each individual module during the experiments in our research building.
The dynamic objects in the scene graph were presented in \cite{lima2023physics}.
1824x16 measurements are used from the Velodyne LiDAR and 640x480 from the Asus Xtion camera; 336384 measurements in total. 
The static parts of the scene consist of 411 698 triangles and 411 698 vertices with a footprint covering an area of 2986.6 square meters.

By exporting the YOLO model as a TensorRT engine, we achieved an average inference time of \SI{15}{\milli\second} per image on the Jetson board, however it consumes a majority of the available GPU resources. Further optimizations include quantization, which allows to use the boards DLAs, which will accelerate inference time and free up GPU resources.
Our proof of concept implementation for the scene graph already results in \SI{1.8}{\hertz} on the Jetson without any further optimization, when doing 4 iterations of ray tracing on the full resolution mesh models (cf.~\autoref{tab:objects}) per frame with 10 Umeyama optimizations for each individual object instance.
Thus it is still being considered fast by the criteria of the BOP Challenge~\cite{hodan2024bop}, with a run time less than \SI{1}{\second} per frame.
The current ray tracing implementation, which takes up the majority of the run time, is limited to Rmagines Embree backend running on the CPU. 
Using Rmagines OptiX backend or Embree with SYCL support should drastically improve performance for ray tracing, as shown in~\cite{mock2023micpl}.

\begin{table}
\caption{The properties of the meshes of the dynamic objects we used in our experiments.}
\label{tab:objects}
\centering

\begin{tabular}{ |c|c|c| }
    \hline
    Object & Faces & Vertices \\ 
    \hline
    multimeter & 100338 & 33446 \\ 
    screwdriver & 26682 & 8894 \\ 
    materialbox & 13617 & 4539 \\ 
    klt3147 & 14124 & 4708 \\ 
    relay & 9312 & 3104 \\
    \hline
\end{tabular}
\vspace{-0.4cm}
\end{table}

\section{Conclusion and Future Work}

In this paper, we presented an approach to detect and track objects in a geometric scene graph.
We connected this front end to a semantic back end to allow spatial reasoning.
Overall, this approach yields good natural tracking results and is expected to significantly reduce the computational cost when tracking multiple objects.
This would be amplified by switching to Rmagine's GPU accelerated backend.
Future work focuses on the efficient realization of spatio-semantic queries utilizing the same underlying acceleration structures and representation, which would allow building and updating hierarchical semantic scene graphs and temporal reasoning in live systems. 
Additionally we plan to evaluate our presented method both on different 6D pose estimation and tracking benchmark data sets, with different initial pose estimators.

\section*{ACKNOWLEDGMENT}
This work is supported by the ExPrIS project through a grant from the German Federal Ministry of Education and Research (BMBF) with Grant Number 01IW23001.

The DFKI Niedersachsen (DFKI NI) is sponsored by the Ministry of Science and Culture of Lower Saxony and the VolkswagenStiftung.

\newpage
\bibliographystyle{IEEEtran}
\bibliography{IEEEabrv, references}

\begin{thebibliography}{10}
\providecommand{\url}[1]{#1}
\csname url@rmstyle\endcsname
\providecommand{\newblock}{\relax}
\providecommand{\bibinfo}[2]{#2}
\providecommand\BIBentrySTDinterwordspacing{\spaceskip=0pt\relax}
\providecommand\BIBentryALTinterwordstretchfactor{4}
\providecommand\BIBentryALTinterwordspacing{\spaceskip=\fontdimen2\font plus
\BIBentryALTinterwordstretchfactor\fontdimen3\font minus
  \fontdimen4\font\relax}
\providecommand\BIBforeignlanguage[2]{{%
\expandafter\ifx\csname l@#1\endcsname\relax
\typeout{** WARNING: IEEEtran.bst: No hyphenation pattern has been}%
\typeout{** loaded for the language `#1'. Using the pattern for}%
\typeout{** the default language instead.}%
\else
\language=\csname l@#1\endcsname
\fi
#2}}

\bibitem{armeni3DSceneGraph2019}
\BIBentryALTinterwordspacing
I.~Armeni, Z.-Y. He, A.~Zamir, J.~Gwak, J.~Malik, M.~Fischer, and S.~Savarese,
  ``{{3D Scene Graph}}: {{A Structure}} for {{Unified Semantics}}, {{3D
  Space}}, and {{Camera}},'' in \emph{2019 {{IEEE}}/{{CVF International
  Conference}} on {{Computer Vision}} ({{ICCV}})}.\hskip 1em plus 0.5em minus
  0.4em\relax IEEE, pp. 5663--5672. [Online]. Available:
  \url{https://ieeexplore.ieee.org/document/9008302/}
\BIBentrySTDinterwordspacing

\bibitem{YoloV8}
\BIBentryALTinterwordspacing
G.~Jocher, A.~Chaurasia, and J.~Qiu, ``Yolo by ultralytics,'' 1 2023. [Online].
  Available: \url{https://github.com/ultralytics/ultralytics}
\BIBentrySTDinterwordspacing

\bibitem{deeken2015semap}
H.~Deeken, T.~Wiemann, K.~Lingemann, and J.~Hertzberg, ``Semap - a semantic
  environment mapping framework,'' in \emph{2015 European Conference on Mobile
  Robots (ECMR)}, 2015, pp. 1--6.

\bibitem{tremblay2018corl:dope}
\BIBentryALTinterwordspacing
J.~Tremblay, T.~To, B.~Sundaralingam, Y.~Xiang, D.~Fox, and S.~Birchfield,
  ``Deep object pose estimation for semantic robotic grasping of household
  objects,'' in \emph{Conference on Robot Learning (CoRL)}, 2018. [Online].
  Available: \url{https://arxiv.org/abs/1809.10790}
\BIBentrySTDinterwordspacing

\bibitem{kehl2016deep}
W.~Kehl, F.~Milletari, F.~Tombari, S.~Ilic, and N.~Navab, ``{Deep learning of
  local rgb-d patches for 3d object detection and 6d pose estimation},'' in
  \emph{Computer Vision--ECCV 2016: 14th European Conference, Amsterdam, The
  Netherlands, October 11-14, 2016, Proceedings, Part III 14}.\hskip 1em plus
  0.5em minus 0.4em\relax Springer, 2016, pp. 205--220.

\bibitem{wang2020self6d}
G.~Wang, F.~Manhardt, J.~Shao, X.~Ji, N.~Navab, and F.~Tombari, ``{Self6d:
  Self-supervised monocular 6d object pose estimation},'' in \emph{Computer
  Vision--ECCV 2020: 16th European Conference, Glasgow, UK, August 23--28,
  2020, Proceedings, Part I 16}.\hskip 1em plus 0.5em minus 0.4em\relax
  Springer, 2020, pp. 108--125.

\bibitem{wu2020eao}
Y.~Wu, Y.~Zhang, D.~Zhu, Y.~Feng, S.~Coleman, and D.~Kerr, ``Eao-slam:
  Monocular semi-dense object slam based on ensemble data association,'' in
  \emph{2020 IEEE/RSJ International Conference on Intelligent Robots and
  Systems (IROS)}.\hskip 1em plus 0.5em minus 0.4em\relax IEEE, 2020, pp.
  4966--4973.

\bibitem{su2022zebrapose}
Y.~Su, M.~Saleh, T.~Fetzer, J.~Rambach, N.~Navab, B.~Busam, D.~Stricker, and
  F.~Tombari, ``Zebrapose: Coarse to fine surface encoding for 6dof object pose
  estimation,'' \emph{arXiv preprint arXiv:2203.09418}, 2022.

\bibitem{Dai_2018_ECCV}
A.~Dai and M.~Niessner, ``3dmv: Joint 3d-multi-view prediction for 3d semantic
  scene segmentation,'' in \emph{Proceedings of the European Conference on
  Computer Vision (ECCV)}, September 2018.

\bibitem{dai2018scancomplete}
A.~Dai, D.~Ritchie, M.~Bokeloh, S.~Reed, J.~Sturm, and M.~Nie{\ss}ner,
  ``Scancomplete: Large-scale scene completion and semantic segmentation for 3d
  scans,'' in \emph{Proceedings of the IEEE Conference on Computer Vision and
  Pattern Recognition}, 2018, pp. 4578--4587.

\bibitem{hou20193d}
J.~Hou, A.~Dai, and M.~Nie{\ss}ner, ``3d-sis: 3d semantic instance segmentation
  of rgb-d scans,'' in \emph{Proceedings of the IEEE/CVF conference on computer
  vision and pattern recognition}, 2019, pp. 4421--4430.

\bibitem{rozenberszki2024unscene3d}
D.~Rozenberszki, O.~Litany, and A.~Dai, ``{Unscene3d: Unsupervised 3d instance
  segmentation for indoor scenes},'' in \emph{Proceedings of the IEEE/CVF
  Conference on Computer Vision and Pattern Recognition}, 2024, pp.
  19\,957--19\,967.

\bibitem{wald2020learning}
J.~Wald, H.~Dhamo, N.~Navab, and F.~Tombari, ``{Learning 3d semantic scene
  graphs from 3d indoor reconstructions},'' in \emph{Proceedings of the
  IEEE/CVF Conference on Computer Vision and Pattern Recognition}, 2020, pp.
  3961--3970.

\bibitem{rosinol2020kimera}
A.~Rosinol, M.~Abate, Y.~Chang, and L.~Carlone, ``{Kimera: an Open-Source
  Library for Real-Time Metric-Semantic Localization and Mapping},'' in
  \emph{2020 IEEE International Conference on Robotics and Automation (ICRA)},
  2020, pp. 1689--1696.

\bibitem{grinvaldVolumetricInstanceAwareSemantic2019}
M.~Grinvald, F.~Furrer, T.~Novkovic, J.~J. Chung, C.~Cadena, R.~Siegwart, and
  J.~Nieto, ``Volumetric {{Instance-Aware Semantic Mapping}} and {{3D Object
  Discovery}},'' vol.~4, no.~3, pp. 3037--3044.

\bibitem{hughes2022hydra}
N.~Hughes, Y.~Chang, and L.~Carlone, ``Hydra: A real-time spatial perception
  system for {3D} scene graph construction and optimization,'' 2022.

\bibitem{hughes2023foundations}
N.~Hughes, Y.~Chang, S.~Hu, R.~Talak, R.~Abdulhai, J.~Strader, and L.~Carlone,
  ``Foundations of spatial perception for robotics: Hierarchical
  representations and real-time systems,'' 2023.

\bibitem{strader2024indoor}
J.~Strader, N.~Hughes, W.~Chen, A.~Speranzon, and L.~Carlone, ``Indoor and
  outdoor 3d scene graph generation via language-enabled spatial ontologies,''
  \emph{IEEE Robotics and Automation Letters}, 2024.

\bibitem{DEEKEN2018146}
\BIBentryALTinterwordspacing
H.~Deeken, T.~Wiemann, and J.~Hertzberg, ``Grounding semantic maps in spatial
  databases,'' \emph{Robotics and Autonomous Systems}, vol. 105, pp. 146--165,
  2018. [Online]. Available:
  \url{https://www.sciencedirect.com/science/article/pii/S0921889017306565}
\BIBentrySTDinterwordspacing

\bibitem{deeken2019spatio}
------, ``A spatio-semantic approach to reasoning about agricultural
  processes,'' \emph{Applied Intelligence}, vol.~49, no.~11, pp. 3821--3833,
  2019.

\bibitem{vizzo2021puma}
I.~Vizzo, X.~Chen, N.~Chebrolu, J.~Behley, and C.~Stachniss, ``{Poisson Surface
  Reconstruction for LiDAR Odometry and Mapping},'' in \emph{International
  Conference on Robotics and Automation (ICRA)}.\hskip 1em plus 0.5em minus
  0.4em\relax IEEE, 2021, pp. 5624--5630.

\bibitem{lin2023immesh}
J.~Lin, C.~Yuan, Y.~Cai, H.~Li, Y.~Ren, Y.~Zou, X.~Hong, and F.~Zhang,
  ``Immesh: An immediate lidar localization and meshing framework,'' \emph{IEEE
  Transactions on Robotics}, vol.~39, no.~6, pp. 4312--4331, 2023.

\bibitem{ruan2023slamesh}
J.~Ruan, B.~Li, Y.~Wang, and Y.~Sun, ``{SLAMesh: Real-time LiDAR Simultaneous
  Localization and Meshing},'' in \emph{International Conference on Robotics
  and Automation (ICRA)}.\hskip 1em plus 0.5em minus 0.4em\relax IEEE, 2023,
  pp. 3546--3552.

\bibitem{mock2023micpl}
A.~Mock, S.~Pütz, T.~Wiemann, and J.~Hertzberg, ``{MICP-L: Mesh ICP for Robot
  Localization using Hardware-Accelerated Ray Casting},'' 2023.

\bibitem{puetz2021cvp}
S.~Pütz, T.~Wiemann, M.~Kleine~Piening, and J.~Hertzberg, ``{Continuous
  Shortest Path Vector Field Navigation on 3D Triangular Meshes for Mobile
  Robots},'' in \emph{International Conference on Robotics and Automation
  (ICRA)}.\hskip 1em plus 0.5em minus 0.4em\relax IEEE, 2021, pp. 2256--2263.

\bibitem{wiemann2018lvr}
T.~Wiemann, I.~Mitschke, A.~Mock, and J.~Hertzberg, ``Surface reconstruction
  from arbitrarily large point clouds,'' in \emph{International Conference on
  Robotic Computing (IRC)}.\hskip 1em plus 0.5em minus 0.4em\relax IEEE, 2018,
  pp. 278--281.

\bibitem{mock2023rmagine}
A.~Mock, T.~Wiemann, and J.~Hertzberg, ``{Rmagine: 3D Range Sensor Simulation
  in Polygonal Maps via Ray Tracing for Embedded Hardware on Mobile Robots},''
  in \emph{International Conference on Robotics and Automation (ICRA)}.\hskip
  1em plus 0.5em minus 0.4em\relax IEEE, 2023, pp. 9076--9082.

\bibitem{embree14}
I.~Wald, S.~Woop, C.~Benthin, G.~S. Johnson, and M.~Ernst, ``{Embree: A Kernel
  Framework for Efficient CPU Ray Tracing},'' \emph{ACM Transactions on
  Graphics (TOG)}, vol.~33, no.~4, jul 2014.

\bibitem{optix10}
S.~G. Parker, J.~Bigler, A.~Dietrich, H.~Friedrich, J.~Hoberock, D.~Luebke,
  D.~McAllister, M.~McGuire, K.~Morley, A.~Robison, and M.~Stich, ``{OptiX: A
  General Purpose Ray Tracing Engine},'' \emph{ACM Transactions on Graphics
  (TOG)}, vol.~29, no.~4, jul 2010.

\bibitem{morrical2021nvisii}
N.~Morrical, J.~Tremblay, Y.~Lin, S.~Tyree, S.~Birchfield, V.~Pascucci, and
  I.~Wald, ``Nvisii: A scriptable tool for photorealistic image generation,''
  2021.

\bibitem{lima2023physics}
O.~Lima, M.~G{\"u}nther, A.~Sung, S.~Stock, M.~Vinci, A.~Smith, J.~Krause, and
  J.~Hertzberg, ``A physics-based simulated robotics testbed for planning and
  acting research,'' in \emph{ICAPS Workshop on Planning and Robotics (PlanRob
  2023)}, 2023.

\bibitem{umeyama1991least}
S.~Umeyama, ``Least-squares estimation of transformation parameters between two
  point patterns,'' \emph{IEEE Transactions on Pattern Analysis \& Machine
  Intelligence}, vol.~13, no.~04, pp. 376--380, 1991.

\bibitem{nieksch2023detecting}
L.~Niecksch, H.~Deeken, and T.~Wiemann, ``{Detecting spatio-temporal Relations
  by Combining a Semantic Map with a Stream Processing Engine},'' in \emph{2023
  IEEE International Conference on Robotics and Automation (ICRA)}, 2023, pp.
  8224--8230.

\bibitem{hodan2024bop}
T.~Hodan, M.~Sundermeyer, Y.~Labbe, V.~N. Nguyen, G.~Wang, E.~Brachmann,
  B.~Drost, V.~Lepetit, C.~Rother, and J.~Matas, ``Bop challenge 2023 on
  detection, segmentation and pose estimation of seen and unseen rigid
  objects,'' 2024.

\end{thebibliography}

\end{document}